\documentclass[runningheads,a4paper]{svproc}
\usepackage[latin1]{inputenc}
\usepackage{amssymb}
\usepackage{graphicx}
\usepackage{amsmath} 
\usepackage{subfigure}

\usepackage{verbatim}
\usepackage{floatrow}
\usepackage{multicol,caption}
\usepackage{multirow}
\newfloatcommand{capbtabbox}{table}[][\FBwidth]

\DeclareCaptionLabelFormat{andtable}{#1~#2  \&  \tablename~\thetable}
\usepackage{lipsum}
\usepackage{etoolbox}

\usepackage{url}
\urldef{\mailsb}\path|{ogil, sanfeliu}@iri.upc.edu|
\newcommand{\keywords}[1]{\par\addvspace\baselineskip
\noindent\keywordname\enspace\ignorespaces#1}

\begin{document}

\mainmatter  

\title{Human motion trajectory prediction using the Social Force Model for real-time and low computational cost applications}

\titlerunning{Human trajectory prediction using the SFM for real-time applications}

\author{\'{O}scar Gil\thanks{Work supported under the European project CANOPIES with grant number H2020- ICT-2020-2-101016906 and JST Moonshot R \& D Grant Number JPMJMS2011-85.}, and Alberto Sanfeliu}

\institute{Institut de Rob\`otica i Inform\`atica Industrial, CSIC-UPC\\
\mailsb}
\authorrunning{}

\maketitle

\vspace{-5mm}

\begin{abstract} 

Human motion trajectory prediction is a very important functionality for human-robot collaboration, specifically in accompanying, guiding, or approaching tasks, but also in social robotics, self-driving vehicles, or security systems. In this paper, a novel trajectory prediction model, Social Force Generative Adversarial Network (SoFGAN), is proposed. SoFGAN uses a Generative Adversarial Network (GAN) and Social Force Model (SFM) to generate different plausible people trajectories reducing collisions in a scene. Furthermore, a Conditional Variational Autoencoder (CVAE) module is added to emphasize the destination learning. We show that our method is more accurate in making predictions in UCY or BIWI datasets than most of the current state-of-the-art models and also reduces collisions in comparison to other approaches. Through real-life experiments, we demonstrate that the model can be used in real-time without GPU's to perform good quality predictions with a low computational cost.

\end{abstract}

\vspace{-7mm}
\section{Introduction}\label{sec_introduction}
Several studies \cite{brown2012role,pezzulo2013action} about Theory of Mind and mirror neurons, emphasize prediction as an essential tool for humans to increase their performance in social interactions through an anticipative behavior. A person can build a model about the internal mental state of people via social interactions to predict future actions.

In particular, human motion prediction is a very broad field with a large number of different categories that depend on multiples factors like the person task, the person model or the person body parts. In the case of human motion trajectory prediction, a very exhaustive taxonomy based on the model approach and the contextual cues have been presented in \cite{rudenko2020human}.

\begin{figure}[bt]
    \centering
    \subfigure{\includegraphics[height=0.2\textwidth]{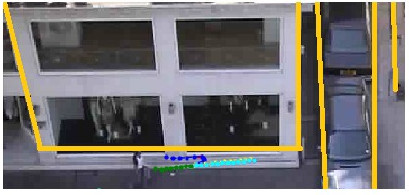}}
    \subfigure{\includegraphics[height=0.2\textwidth]{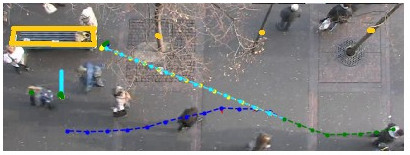}}
    \caption{{\bf Complex cases.} The left picture shows a person who randomly changes the movement direction because is waiting for someone. The right picture shows a bench that can be an obstacle for some pedestrians or a goal for others.}
\vspace{0mm}
\label{paper_resume}
\end{figure}

Human motion trajectory prediction is a complex task very difficult to understand, due to the very different strategies people use to avoid collisions; the variety of social interactions; the relative nature of consider something an obstacle or a goal; and the sudden changes in the movements due to internal unpredictable stimulus (refer to Fig.\ref{paper_resume}). 

The stimuli for pedestrian motion can be internal or external. The internal stimuli are very difficult to infer because they are related to the particular person's thoughts. The external stimuli are related to the environment, but the response of the stimulus is related to the person's mental state as well. On the whole, any response of the person depends on external and internal stimuli and there aren't handwritten rules or laws to explain all the cases. 

Due to this complex dependence, this work uses the environment information through the Social Force Model (SFM) \cite{helbing1995} and people features like the velocities and the resultant forces. This information is used to generate a set of possible paths. The advantage of generating a set of paths is that the multimodal behavior of pedestrians can be handled.

The remainder of this paper is organized as follows. In section 2, the related work is introduced. In Section 3, the SFM used to encode the environment information is described. In Section 4, a description of the complete approach, that combines a Generative Adversarial Network (GAN) and a Conditional Variational Autoencoder (CVAE) is introduced. Section 5 provides an analysis of the metrics to evaluate the models and make state-of-the-art comparison as fair as possible. In Section 6, the evaluation by the usual methodology in different datasets is performed and the real-life experiment results are analyzed. Finally, in section 7, the conclusions are provided.

\section{Related Work}

Nowadays, multiple approaches to human motion prediction have been developed. These models are normally data-driven models to forecast the next skeleton 3D movements in concrete tasks like, for example, walking, eating, smoking, or running \cite{martinez2017human,li2020exploring}. Commonly, these data-driven methods formulate the problem as a sequence-to-sequence task where the data is processed by a Recurrent Neural Network (RNN) or a Long Short-Term Memory (LSTM). These network architectures are chosen due to their ability to encode temporal information.

In human motion trajectory prediction, as a part of human motion prediction, there are a lot of elements in common like social interaction or multimodal behavior. By contrast, there are fewer dependencies on the task, and the person model can be simplified, for example by considering a point or 2D circle.

Physics-based approaches, like Constant Velocity Model (CVM) \cite{scholler2020constant} or Extended Kalman Filter (EKF), are simpler than other methods, but can give a good performance, for example, in linear trajectories. These approaches can be obstacle-aware methods like the Social Force Model \cite{helbing1995}, which can be combined with goal estimation \cite{ferrer2014bayesian} for better performance.

Actually, data-driven models are very common in trajectory prediction. Most of these approaches utilize LSTM architectures. In this paragraph, we discuss  models that do not take into account static obstacles or collisions. A first method is Social LSTM \cite{alahi2016}, which uses an encoder-decoder structure with LSTM cells and a pooling mechanism to encode the people that is close to a person. The MX-LSTM model \cite{hasan2018} considers not only people positions (tracklets), but gaze direction (vislets) too. A transformer is proposed in the AgentFormer model \cite{yuan2021agent}, which learns simultaneously the social and time dimensions. Social GAN \cite{gupta2018} uses a GAN and a pooling mechanism to generate a multimodal distribution of the trajectories. In PECNet \cite{mangalam2020not}, a CVAE is included to obtain an accurate estimation of the goals.

Furthermore, obstacle-aware models like Next \cite{liang2019peeking} and SoPhie \cite{sadeghian2019sophie} extract features from the image of the scene using a convolutional neural network (CNN) and then, a RNN is utilized to obtain predictions. Other works, like Trajectron$++$ \cite{salzmann2020trajectron++}, use a CNN to incorporate the map information and take into account the dynamics through an RNN. NSP model \cite{Jiang_trajectory_2022} uses the Social Force Model (SFM) and a Neural Differential Equation for the motion prediction. In \cite{mangalam2021goals}, the scene features are used to obtain waypoints along the trajectory to improve the results in a long prediction horizon.

\section{Social Force Model}

The Social Force Model of Helbing and Molnar \cite{helbing1995} for pedestrian dynamics allows to simulate social interactions as forces. In a scene with a set of pedestrians $P$ and obstacles $O$, it is considered that a pedestrian $ p \in P$ moves towards a goal with the following attractive force:
\begin{equation} {\bf f^{goal}_{p}} = k({\bf v^{0}_{p}}-{\bf v_{p}})
\label{eq1}
\end{equation}
where ${\bf v_{p}}$  is the current velocity of the pedestrian and $k^{-1}$ is the relaxation time to achieve the desire velocity pointing towards the goal, ${\bf v^{0}_{p}}$.

To consider the influence of pedestrians and obstacles, avoid collisions and respect social distances, a repulsive force is defined as: 
\begin{equation} {\bf f_{z,p}^{int}}=A_{z}e^{\left(d_{z}-d_{z,p}\right)/B_{z}} {\bf \hat{d}_{z,p}} 
\label{eq2}
\end{equation}
where $z$ can be a pedestrian or obstacle. $A_{z}$, $B_{z}$ and $d_{z}$ are parameters that can be adjusted. $d_{z,p}$, is the Euclidean distance between $z$ and the pedestrian $p$. ${\bf \hat{d}_{z,p}}$ is the unitary vector in the line between $p$ and $z$ positions, pointing to $p$.

Zanlungo et al. \cite{zanlungo2011} consider collision prediction into the interaction force between pedestrians $p,q \in P$. In this case the repulsive force for pedestrian $p$ is:

\begin{equation} {\bf f_{q,p}^{int}}(\{ {\bf v_{q,p}} \}, \{ {\bf d_{q,p}} \}, {\bf v_{p}})=A_{q}\frac{v_{p}}{t_{p}}e^{-d_{q,p}/B_{q}} \frac{{\bf d'_{q,p}(t_{p})}}{d'_{q,p}(t_{p})}
\label{eq3}
\end{equation}
where $\{ {\bf v_{q,p}} \}$ is the set of relative velocities between $p$ and other pedestrians. $\{ {\bf d_{q,p}} \}$ is the set of vectors with all relative distances between $p$ and other pedestrians, pointing to $p$. $t_{p}$=min$_{q}\{ t_{q,p} \}$, where $t_{q,p}$ is the time in which $p$ is at the minimum distance from $q$ and ${\bf d'_{q,p}(t_{p})}$ is the relative position of $p$ regarding $q$, in $t$=$t_{q,p}$. When the angle between ${\bf v_{q,p}}$ and ${\bf d_{q,p}}$ complies with $\left| \theta_{p,q} \right|> \pi/4$, then $t_{q,p}=\infty$. Bearing in mind these forces, the resulting force for a pedestrian $p$ is:
\begin{equation} {\bf F_{p}}={\bf f^{goal}_{p}}+\sum_{q \in P} {\bf f_{q,p}^{int}}+\sum_{o \in O} {\bf f_{o,p}^{int}}
\label{eq4}
\end{equation}
This model has been generalized to robots by means of the Extended Social Force Model (ESFM). This generalization is very useful for planning in collaborative tasks that involve social-aware navigation \cite{repiso2019adaptive,repiso2018robot}.\\

\section{Social Force GAN Model}
\label{sec:4}

In this section, we describe approach which considers SFM, a GAN and a CVAE. 

\subsection{Problem Formulation}
Trajectory prediction in an environment can be considered as a prediction of multiple positions of points along different discrete times (timesteps). In order to make predictions, an observation horizon, $T_{obs}$ and a prediction horizon of timesteps, $T_{pred}$, are established in a trajectory ${\bf X_{i}}$, for a pedestrian $i$, with the correspondent ground truth positions $(x_{i}^{t},y_{i}^{t}) \in {\bf X_{i}}$ associated in each horizon: 
\begin{equation} {\bf X_{i}^{obs}}=\{(x_{i}^{t},y_{i}^{t}) | t=1,2, \ldots ,T_{obs} \} 
\label{eq5}
\end{equation}
\begin{equation} {\bf X_{i}^{pred}}=\{(x_{i}^{t},y_{i}^{t}) | t=T_{obs}+1, \ldots ,T_{obs}+T_{pred} \} 
\label{eq6}
\end{equation}
The main objective is to forecast the future positions of the trajectory:
\begin{equation}{\bf Y_{i}}=\{(x_{i}^{t},y_{i}^{t}) | t=T_{obs}+1, \ldots ,T_{obs}+T_{pred} \}
\label{eq7}
\end{equation}
as close as possible to the ground truth positions in the prediction horizon, ${\bf X_{i}^{pred}}$. To achieve this goal, the previous positions of the trajectory, ${\bf X_{i}^{obs}}$, can be used as information. Other cues as the gaze, the potential goals, the neighbors' positions or the environment map can also be used.

\subsection{Social Force Representation}

SFM is used in this work to obtain an environment representation $R$, as a part of the inputs for the Social Force GAN model (SoFGAN). Given a pedestrian $p \in P$, the environment representation for this pedestrian is calculated using (\ref{eq2}), to calculate repulsive forces because of static obstacles, and using (\ref{eq3}) to calculate forces due to other pedestrians.

These 2 types of repulsive forces could be added separately or in 1 unique force but, in the last case, the resultant forces would not give a complete description of the environment, because there are a lot of different combinations of static obstacles or pedestrians that can give the same resultant forces.  

\begin{figure}[bt]
    \centering
    \includegraphics[width=0.5\linewidth]{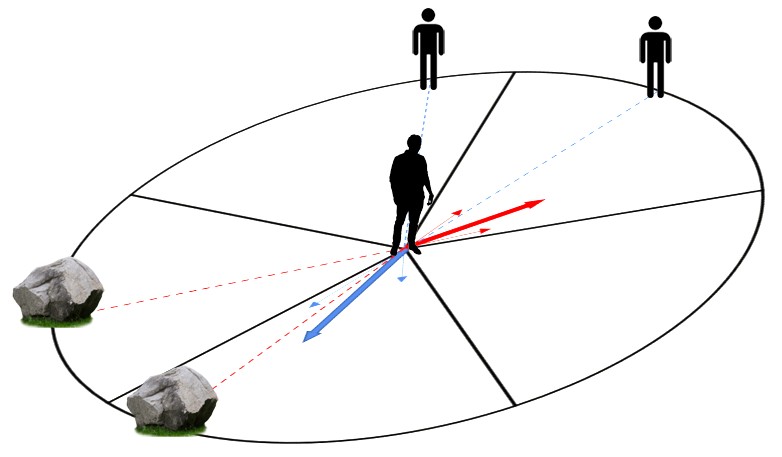}
    \caption{{\bf Social Force Representation.} In this example 5 angle bins are considered. The thick vectors are the resultant forces of the thin vectors in each angle bin. There are 2 blue forces of 2 pedestrians and 2 red forces of 2 obstacles applied in a pedestrian located in the center.}
\vspace{0mm}
\label{bins}
\end{figure}

Therefore, to avoid this problem, $M$ angle bins centered in pedestrian $p$ are considered to divide the space. Using this method, two resultant forces are calculated for each angle bin. One for the static obstacles and another for the pedestrians. The two sets of forces are used as the Social Force Representation:
\begin{equation}R_{p}=\{ {\bf F_{ped}^{bin}}, {\bf F_{obst}^{bin}}\}
\label{eq8}
\end{equation}
where each set of forces is:
\begin{equation}{\bf F_{ped}^{bin}}=\{ \sum_{q \in P_{i}} {\bf f_{q,p}^{int}}\}_{i=1}^{M} 
\label{eq9}
\end{equation}
\begin{equation}{\bf F_{obst}^{bin}}=\{ \sum_{o \in O_{i}} {\bf f_{o,p}^{int}}\}_{i=1}^{M} 
\label{eq10}
\end{equation}
$O_{i}$ and $P_{i}$ are the subsets of obstacles and pedestrians whose forces applied in $p$ are into the angle bin $i$ and comply with $\bigcup_{i=1}^{M}O_{i}=O$ and $\bigcup_{i=1}^{M}P_{i}=P$. An example is shown in Fig. \ref{bins}. 

The forces of the static obstacles are calculated modeling obstacles in the environment as polygons and choosing always the nearest point of the polygon to $p$ as the obstacle point to calculate the force. The homography matrix of the images is used to obtain the Cartesian coordinates of the polygons.
Taking into account this representation, the inputs of the SoFGAN model for all timesteps in the observation horizon are as follows:
\begin{itemize}
\item ${\bf f_{p}^{goal}}$: The attractive force calculated using (\ref{eq1}) for each pedestrian.
\item $R_{p}$: The Social Force Representation.
\item $ {\bf X_{rel}^{obs}} $: The relative positions of each pedestrian in the observation horizon and considering the first relative position zero.
\item ${\bf F_{total}^{obs}}$: The necessary forces at each timestep to generate the trajectory starting at the first position.
\item ${\bf X_{r}^{obs}}$: The relative positions of each pedestrian in the observation horizon taking as reference the first position in the trajectory.
\item ${\bf g_{r}^{GT}}$: The final ground truth position for each trajectory taking as reference the first position in the trajectory. Only used during training.
\end{itemize}

The forces to generate the trajectory, ${\bf F_{total}^{obs}}$, are calculated using the relative positions and the timestep value (to compute their velocities). Considering $\Delta t=1$ and a unitary mass, the force is calculated as a difference of the straight sections in the trajectory at each timestep. For the attractive force, the goal is in the last position of the observation horizon.\\

\subsection{SoFGAN model description}

The model can be summarized as:
\begin{equation} \{{\bf Y_{rel}},{\bf F_{total}^{pred}}\}=SoFGAN({\bf X_{rel}^{obs}}, {\bf f_{p}^{goal}}, R_{p}, {\bf F_{total}^{obs}}, {\bf X_{r}^{obs}}, {\bf g_{r}^{GT}}) 
\label{eq11}
\end{equation}
where ${\bf Y_{rel}}$ is the set of the predicted relative positions from which the absolute positions for all pedestrians $\{{\bf Y_{i}}\}_{i=1}^{N}$ can be calculated. ${\bf F_{total}^{pred}}$ are the necessary forces to generate at each timestep a new position, to form a second predicted trajectory. Therefore, SoFGAN can also be considered as a force predictor. 

\begin{figure*}[t!]
    \centering
    \includegraphics[width=1\linewidth]{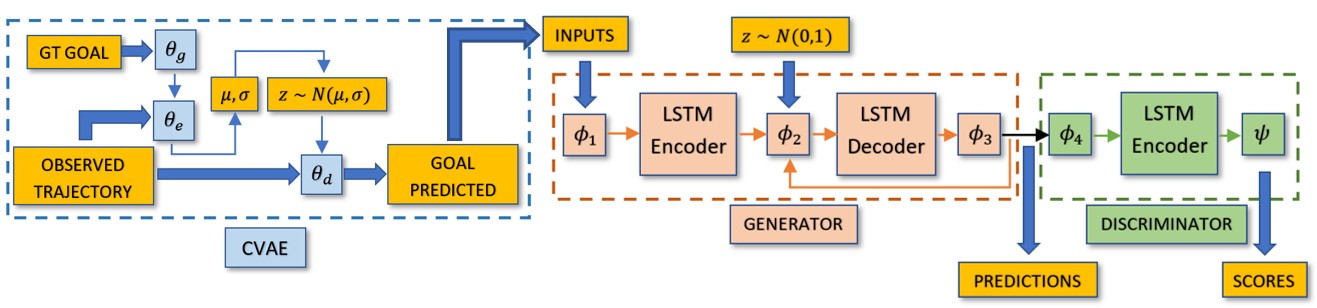}
    \caption{{\bf Social Force GAN with CVAE architecture.} The GAN generator provides the predictions using the predicted goals given by the CVAE module.}
\vspace{0mm}
\label{model}
\end{figure*}

The architecture of the model is basically a GAN, like in \cite{gupta2018} and a CVAE module. The complete model is shown in Fig. \ref{model}, where $\phi_{1}$, $\phi_{2}$, $\phi_{3}$ and $\phi_{4}$ are linear transformations. $\psi$, $\theta_{g}$, $\theta_{e}$, $\theta_{d}$ are MLP's. 

The CVAE module is used to reduce the main source of the prediction errors, the last position prediction. For that reason, it is used to estimate the goal trajectories. The inputs for the CVAE are ${\bf X_{r}^{obs}}$ and ${\bf g_{r}^{GT}}$. The second one is only used during training. During the test, $\theta_{g}$ and $\theta_{e}$ are not used. The only necessary input for the CVAE in the test phase is the observed trajectory because ${\bf z}$ is sampled from $N({\bf 0},\alpha{\bf I})$, a normal distribution with mean zero and a fixed variance $\alpha$. If the CVAE module is not used, the GAN inputs are ${\bf X_{rel}^{obs}}$, ${\bf f_{p}^{goal}}$, $R_{p}$ and ${\bf F_{total}^{obs}}$. In case of use the CVAE module, the attractive force is substituted by the CVAE predicted goals, ${\bf g_{r}^{pred}}$, because it improves the model results.

The scores obtained in the discriminator are used as labels to train the model. The losses used to train the GAN and the CVAE in this approach are the adversarial loss, the variety loss \cite{gupta2018} and the CVAE loss. The CVAE loss is composed by the Kullback-Leibler (KL) divergence and the variety loss applied to the last predicted position:
\begin{equation}
    L_{adv}= \min_{G} \max_{D} [\mathbb{E}_{x \sim p_{d}} log D_{\theta_{d}} (x) + \mathbb{E}_{z \sim p(z)} log(1-D_{\theta_{d}} (G_{\theta_{g}} (z)))]
\label{eq21}
\end{equation}
\begin{equation}
    L_{var}= \min_{k} \|{\bf Y_{rel}^{k}}-{\bf X_{rel}^{k,pred}}\|
\label{eq22}
\end{equation}
\begin{equation}
L_{CVAE}= \lambda_{2}D_{KL}(N(\boldsymbol{\mu},\boldsymbol{\sigma}) \|N({\bf 0},{\bf I}))+\lambda_{3}\min_{k} \|{\bf g_{r}^{pred}}-{\bf g_{r}^{GT}}\|
\label{eq29}
\end{equation}
\begin{equation}
    L_{total}=\mathbb{E}_{p \in P}[L_{adv}+\lambda_{1} L_{var}+L_{CVAE}]
\label{eq30}
\end{equation}
where $k$ is the number of generator samples. 

\section{Metrics}
\label{sec:6}

In this work, as in \cite{yuan2021agent} and \cite{salzmann2020trajectron++}, $k$ trajectory samples are used to compute the minimum Average Displacement Error (mADE) and the minimum Final Displacement Error (mFDE) separately over the sampled scenes that contain different number of pedestrians. 

Another metric to measure the social-awareness of the model is the average \% of colliding pedestrians per frame in a dataset, $\%c$. This value is used in different works like \cite{sadeghian2019sophie}, although they do not give a detailed explanation about how to obtain this metric. In this work, a collision is detected when 2 pedestrians get closer than $0.1 \ m$ in a frame, but not between successive frames. The $\%c$ of collisions is then computed as follows:
\begin{equation}
    \%c =\sum_{j=1}^{k}\frac{1}{k} \left( \sum_{i=1}^{N_{f}}\frac{1}{N_{f}}(\frac{100N^{c}_{ij}}{N_{ij}}) \right)
\label{eq39}
\end{equation}
where $N_{f}$ is the number of predicted timesteps in the test set, $k$ is the number of generator samples, $N_{ij}^{c}$ is the number of colliding pedestrians in timestep or frame $i$ in the sample $j$. $N_{ij}$ is the total number of pedestrians in the timestep $i$ and sample $j$.

In this work, as usual, the observation horizon is $T_{obs}=3.2 s$, which corresponds to 8 timesteps. The prediction horizon is $T_{pred}=4.8 s$, which corresponds to 12 timesteps.

\section{Experiments}

The SoFGAN model has been evaluated through the BIWI \cite{pellegrini2009you}, UCY \cite{lerner2007crowds} datasets using the leave-one-out cross validation technique as in \cite{alahi2016}, \cite{gupta2018} and \cite{sadeghian2019sophie}. Moreover, the model has been implemented in ROS to predict people in real-time with a low computational cost.

\subsection{Evaluation Results}

\begin{table*}[!h]
\begin{center}
\resizebox{\columnwidth}{!}{%
\begin{tabular}{c c c c c c c}
\hline
Model & {\bf ETH} & {\bf HOTEL} & {\bf ZARA1} & {\bf ZARA2} & {\bf UNIV} & {\bf AVG}  \\
\hline \hline 

{\bf SGAN \cite{gupta2018}}  & 0.81/1.52 & 0.72/1.61 & 0.34/0.69 & 0.42/0.84 & 0.60/1.26 & 0.58/1.18 \\

{\bf SoPhie \cite{sadeghian2019sophie}}  & 0.70/1.43 & 0.76/1.67 & 0.30/0.63 & 0.38/0.78 & 0.54/1.24 & 0.54/1.15 \\

{\bf Next \cite{liang2019peeking}}  & 0.73/1.65 & 0.30/0.59 & 0.38/0.81 & 0.31/0.68 & 0.60/1.27 & 0.46/1.00 \\

{\bf CVM-20 \cite{scholler2020constant}} & 0.96/2.09 & 0.29/0.54 & 0.52/1.03 & 0.34/0.70 & 0.61/1.26 & 0.54/1.12\\

{\bf PECNet \cite{mangalam2020not}} & 0.54/0.87 & 0.18/0.24 & 0.22/0.39 & 0.17/0.30 & 0.35/0.60 & 0.29/0.48\\

{\bf Trajectron++ \cite{salzmann2020trajectron++}} & 0.57/1.05 & 0.16/0.26 & 0.22/0.41 & 0.16/0.31 & 0.28/0.56 & 0.28/0.52\\

{\bf AgentFormer \cite{yuan2021agent}} & 0.45/0.75 & 0.14/0.22 & 0.18/0.30 & 0.14/0.24 & 0.25/0.45 & 0.23/0.39\\

{\bf Y-NET \cite{mangalam2021goals}} & 0.28/0.33 & 0.10/0.14 & 0.17/0.27 & 0.13/0.22 & 0.24/0.41 & 0.18/0.27\\

{\bf NSP \cite{Jiang_trajectory_2022}} & {\bf 0.25/0.24} & {\bf 0.09/0.13} & {\bf 0.16/0.27} & {\bf 0.12/0.20} & {\bf 0.21/0.38} & {\bf 0.17/0.24}\\

\hline
{\bf SoFGAN} & 0.44/0.68 & 0.13/0.18 & 0.20/0.35 & 0.17/0.31 & 0.25/0.46 & 0.24/0.40\\


{\bf fSoFGAN} & 0.48/0.79 & 0.15/0.22 & 0.22/0.42 & 0.20/0.38 & 0.28/0.53 & 0.27/0.47\\
\hline

\hline 
\end{tabular}
}
\end{center}
\caption{{\bf ${\bf mADE/mFDE}$ results.} Comparison between SoFGAN and other models for 20 samples. Numbers in bold type are the best results.}
\label{metric_a}
\end{table*}

The results of the evaluation in terms of $mADE$ and $mFDE$, for BIWI and UCY datasets, are shown in Table \ref{metric_a}, where SoFGAN is the model with a CVAE trained using data augmentation. To improve the performance, the 20 samples are selected through a k-means clustering of 1000 samples, as in \cite{mangalam2021goals}. fSoFGAN evaluate the $mADE$ and $mFDE$ of the generated trajectories using the total forces. For SGAN \cite{gupta2018}, SoPhie \cite{sadeghian2019sophie} and Next \cite{liang2019peeking} the paper results in \cite{liang2019peeking} are exposed. CMV-20 use 20 samples and it has been implemented because the evaluation in \cite{scholler2020constant} is different.


The Trajectron++ results are different from the paper because an error in the velocity and acceleration estimation has been corrected. The Y-NET and NSP results have been obtained from the papers directly because the datasets and hyperparameters to reproduce the evaluation are not public. Only the AgentFormer model is the one that has been tested obtaining slightly better results than the SoFGAN model. 

Table \ref{colisions} shows the $\%c$ for different models where GT is the ground truth. It is important to underscore that the results of SGAN and SoPhie $\%c$ are the ones that appear in their corresponding papers, although we do not know if their $\%c$ calculation method is the same as our method. The same occurs with NSP and the Y-NET $\%c$. Nevertheless, the improvement compared with CVM-20 is very significant. Although SoFGAN does not provide the best results in terms of mADE, mFDE and $\%c$, it outperforms most of state of the art methods. The SoFGAN and NSP models show that the use of forces can be very useful to encode the environment information and improve the predictions.

\begin{table}[h]
\begin{center}
\begin{tabular}{c c c c c c c}
\hline
Model &{\bf ETH} & {\bf HOTEL} & {\bf ZARA1} & {\bf ZARA2} & {\bf UNIV} & {\bf AVG}\\
\hline \hline 
{\bf GT}  & 0.000 & 0.000 & 0.000 & 0.000 & 0.056 & 0.011 \\

{\bf SGAN \cite{gupta2018}}  & 2.509 & 1.752 & 1.749 & 2.020 & {\bf 0.559} & 1.717\\

{\bf SoPhie \cite{sadeghian2019sophie}}  & 1.757 & 1.936 & 1.027 & 1.464 & 0.621 & 1.361 \\

{\bf CVM-20 \cite{scholler2020constant}}  & 1.764 & 1.430 & 2.680 & 2.163 & 4.172 & 2.442 \\

{\bf Y-NET \cite{mangalam2021goals}}  & {\bf 0.000} & {\bf 0.000} & 0.820 & 1.310 & 1.510 & 0.730 \\

{\bf NSP \cite{Jiang_trajectory_2022}}  & {\bf 0.000} & {\bf 0.000} & {\bf 0.000} & {\bf 0.660} & 1.480 & {\bf 0.430} \\
\hline
{\bf SoFGAN}  & 0.250 & 0.500 & 0.707 & 1.226 & 3.688 & 1.274 \\

\hline 
\end{tabular}
\end{center}
\caption{{\bf \%c across BIWI and UCY datasets for 20 samples.}}
\label{colisions}
\end{table}

An ablation study has been performed to demonstrate the benefits of the CVAE and the data augmentation. M(w/o1) is the model without the CVAE. M(w/o2) is the M(w/o1) model without data augmentation. The average mADE and mFDE are shown in Table \ref{ablation} for the BIWI and UCY datasets.

\begin{table*}[!h]
\begin{center}
\begin{tabular}{c c c c}
\hline
Model & {\bf SoFGAN} & {\bf M(w/o1)} & {\bf M(w/o2)}  \\
\hline \hline 

{\bf AVG}  & {\bf 0.24/0.40} & 0.25/0.44 & 0.29/0.52 \\


\hline 
\end{tabular}

\end{center}
\caption{{\bf ${\bf mADE/mFDE}$ results of the ablation study.}}
\label{ablation}
\end{table*}

\subsection{Real-life Experiments}

Through a ROS implementation in the Helena robot, real outdoor experiments have been performed with people. Helena is a transporter robot with a RS-LiDAR-16 and a Pioneer P3-DX. Although the Helena prediction is not computed, Helena is taken into account as an agent to compute the forces.

\begin{figure*}[t!]
    \centering
    \subfigure[RViz]{\includegraphics[height=0.16\textwidth]{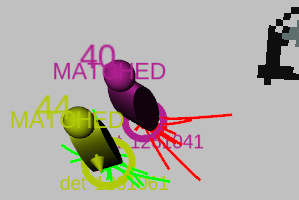}}
    \subfigure[RViz]{\includegraphics[height=0.16\textwidth]{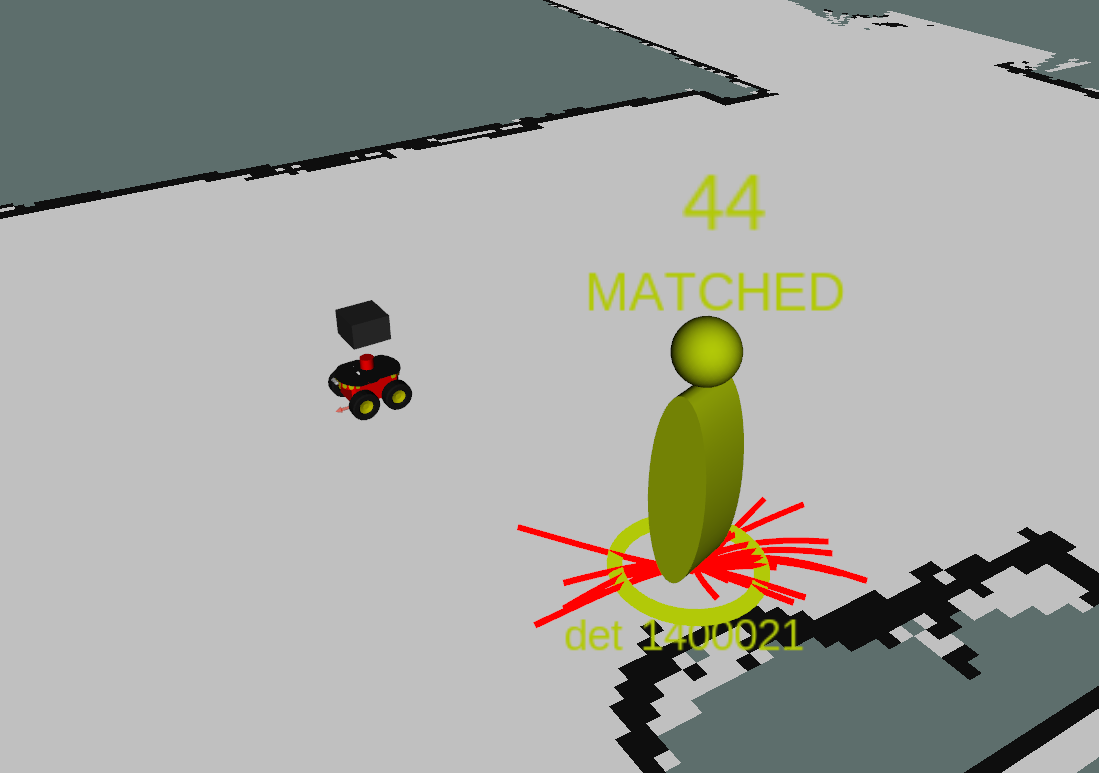}}
    \subfigure[Real World]{\includegraphics[height=0.16\textwidth]{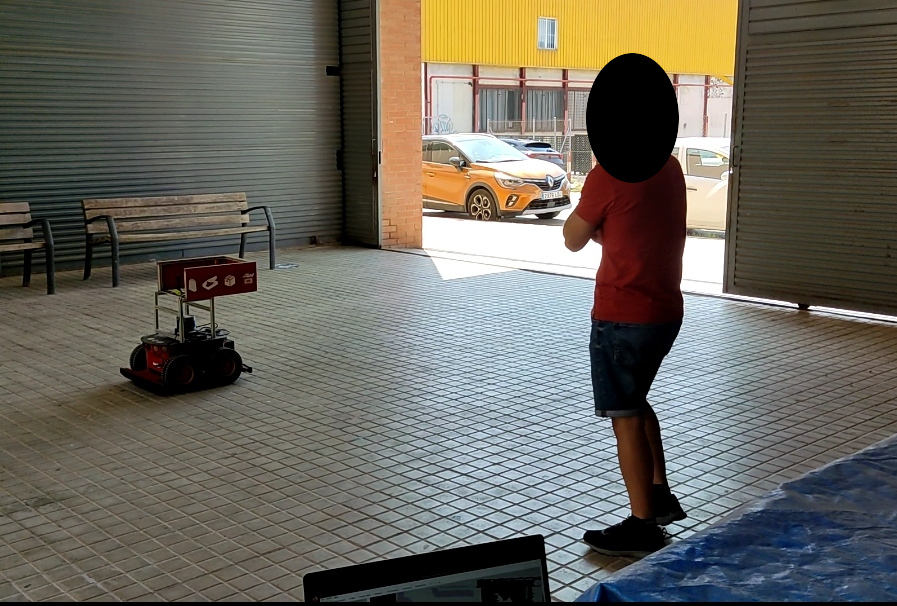}}
    \subfigure[RViz]{\includegraphics[height=0.16\textwidth]{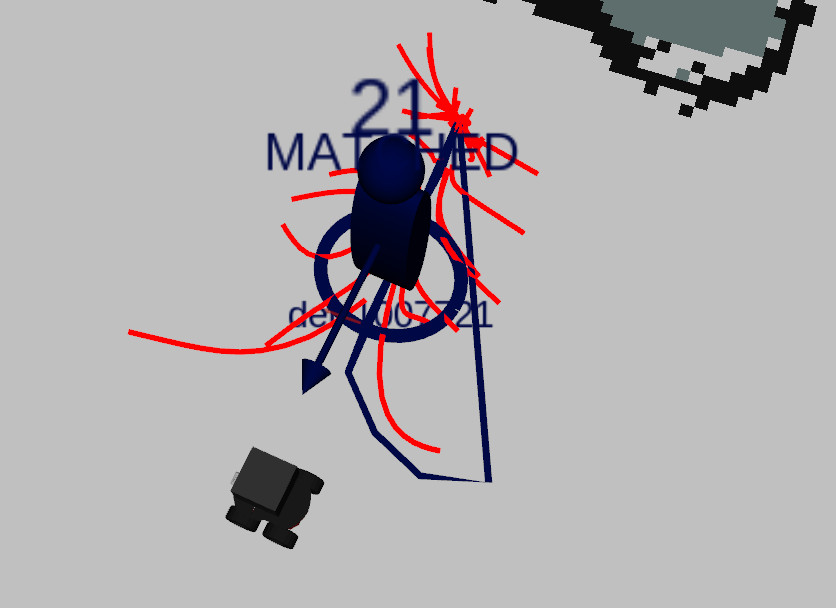}}
    \caption{{\bf Predictions when Helena is not moving.} The predictions are the red and green lines in the ground. The visualization is obtained using RViz.}
\vspace{0mm}
\label{exp_1}
\end{figure*}

The first experiment is to evaluate the predictions when the Helena robot is not moving. In Fig. \ref{exp_1}, from left to right, in the first image the interaction between 2 people in a conversation is shown. The model does not sample trajectories between them. In the second image, the effect of the obstacle behind the person cause that most of predictions appear in different directions. In the last image, the person walks towards Helena. In this case, the predictions take into account the robot and try to avoid collisions.

\begin{figure*}[t!]
    \centering
    \subfigure[RViz: t]{\includegraphics[height=0.18\textwidth]{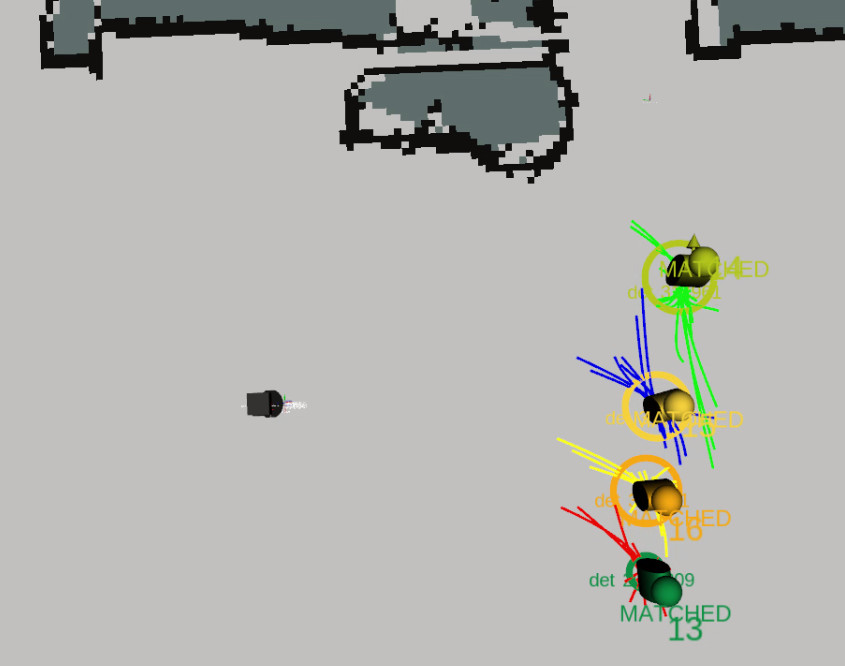}}
    \subfigure[RViz: t+1]{\includegraphics[height=0.18\textwidth]{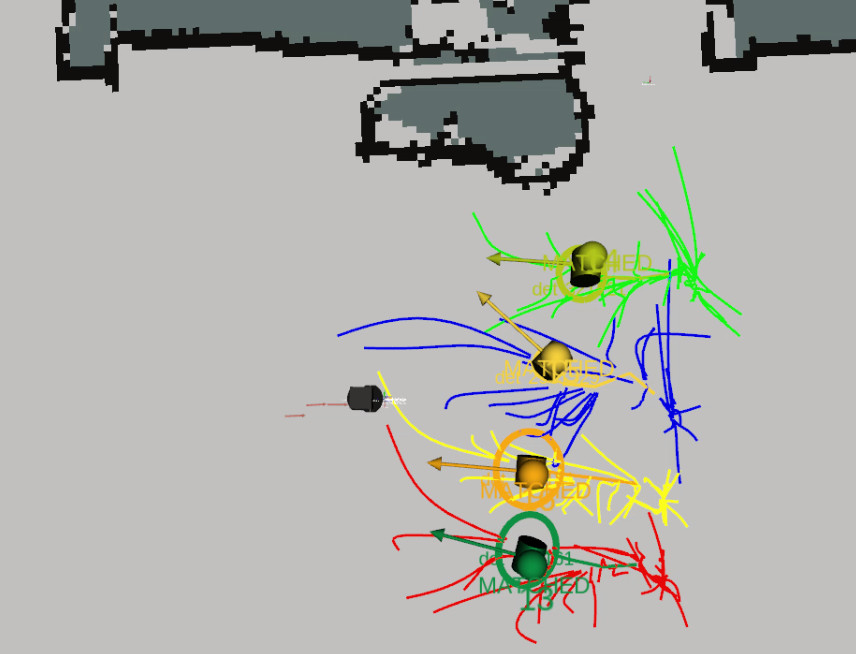}}
    \subfigure[RViz: t + 2]{\includegraphics[height=0.18\textwidth]{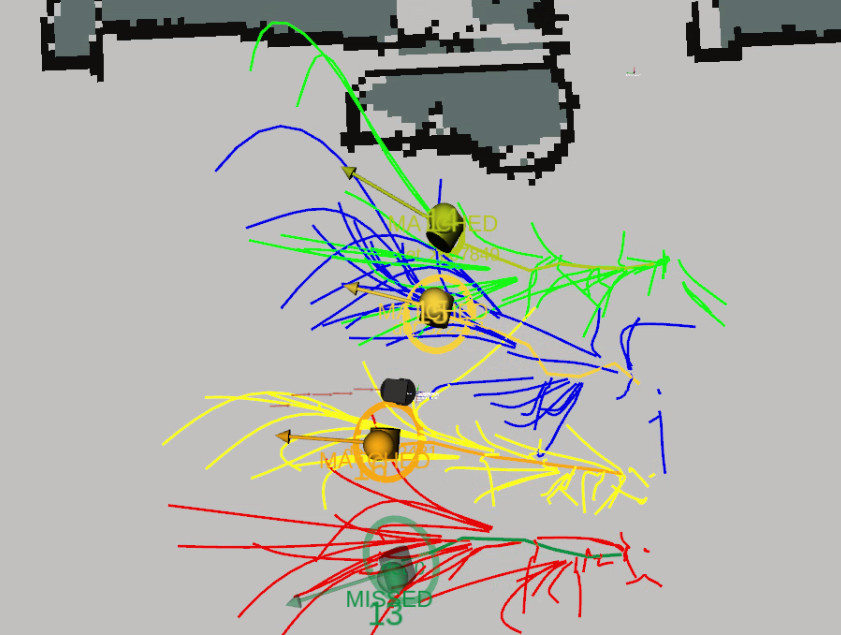}}
    \subfigure[Real World]{\includegraphics[height=0.18\textwidth]{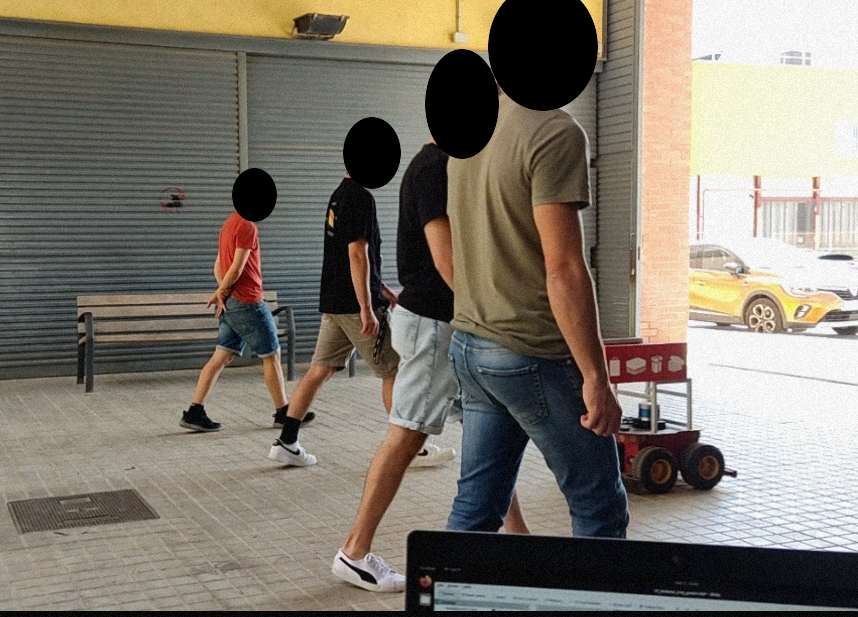}}
    \caption{{\bf Predictions during an encounter between 4 people and Helena.} The predictions are the colored lines in the ground.}
\vspace{0mm}
\label{exp_3}
\end{figure*}

The second experiment, shown in Fig. \ref{exp_3}, evaluates the interaction between 4 people and the interaction between the group and the robot. From left to right, in the first image, the group is not moving and the predictions are short and avoid collisions between them. In the second image, the group moves and the predictions try to avoid collisions with the robot. In the third image, the people change their velocity directions due to the robot and the blue and green predictions close to obstacles try to avoid a collision.

\subsection{Implementation Details}

For this work we have used the Adam optimizer, with a learning rate of 0.0005 for both generator and discriminator. For the variety loss, the $\lambda_{1}$ weight is 0.5 and the number of samples is 20. When the CVAE module is added, $\lambda_{1}$ is the same, $\lambda_{2}$ is 1 and $\lambda_{3}$ is 0.5. The CVAE $\mu$ parameter for the probability distribution has 16 components and $\boldsymbol{\sigma} = \alpha {\bf I}$ is a $16 \times 16$ matrix. For testing $\mu$ is zero and $\alpha$ is 3. The parameters for $\mu$ and $\sigma$ have been chosen between other combinations to obtain a good performance.  For noise, the dimension is 32 and the dimensions for hidden states are 64 for each encoder and 128 for the decoder. The embedding dimension is 64 and the batch size is 256.

The LSTM encoder and decoder have one layer. The MLP, $\psi$, used in the discriminator, has dimensions (64, 1024, 1). The CVAE MLP's, $\theta_{g}$, $\theta_{e}$, $\theta_{d}$ have these dimensions successively: (2, 8, 16, 16), (32, 8, 50, 32) and (32, 1024, 512, 1024, 2). All the layers use ReLu as activation function and batch normalization, except for the last layer of $\theta_{d}$.

To calculate the Social Force Representation, four angle bins have been considered. All the models have been trained using Pytorch in a Tesla K40 GPU.

The real experiments have been performed using a CPU and ROS Melodic. The Lidar measures are provided to the Spencer tracker \cite{linder2016multi} to detect moving obstacles. The static obstacles are considered using the 2-D occupancy grid of the environment map. This map is used for robot localization and navigation in ROS. The model can compute the predictions in less than $100 \ ms$.

\section{Conclusions}
In this work, we have presented a new human trajectory predictor model denominated Social Force Generative Adversarial Network (SoFGAN). This new trajectory predictor, is based on Social Force Model (SFM) and Generative Adversarial Network (GAN). One of the main advantages of this new trajectory predictor, is that includes the social forces of the environment and the moving pedestrians, to improve the human prediction in the next timesteps. Additionally, we have included a CVAE in order to learn the trajectory goal distribution resulting in an improvement of the model. The experimental results on standard datasets of complex human motions, show that our predictor gets good results in comparison to the best state of the art methods that we have tested. Moreover, we also obtain good results in the percent of colliding pedestrians per frame, $\%c$, in ETH/UCY datasets. Unlike other authors, we demonstrate that our model can be used in real time applications with a low computational cost to perform human-like predictions.


%
%

\bibliographystyle{splncs03}

\end{document}